\documentclass[preprint,12pt]{elsarticle}

\usepackage{caption}
\usepackage{graphicx}
\usepackage{amssymb}
\usepackage{siunitx}
\usepackage{amsmath}
\usepackage{subfig}

\journal{arXiv}

\begin{document}

\begin{frontmatter}


\title{Weight-Based Exploration for Unmanned Aerial Teams Searching for Multiple Survivors}



\author[label1]{Sarthak J. Shetty\corref{cor1}}
\author[label1]{Debasish Ghose\corref{cor1}}
\cortext[cor1]{Sarthak J. Shetty is a Research Associate and Debasish Ghose is a Professor in the Guidance, Control, and Decision Systems Laboratory (GCDSL), Department of Aerospace Engineering, Indian Institute of Science, Bangalore, India. \\
E-mail: \texttt{<sarthakjs + dghose>@iisc.ac.in}}
\address[label1]{Department of Aerospace Engineering, Indian Institute of Science, Bengaluru, India}

\begin{abstract}
During floods, reaching survivors in the shortest possible time is a priority for rescue teams. Given their ability to explore difficult terrain in short spans of time, Unmanned Aerial Vehicles (UAVs) have become an increasingly valuable aid to search and rescue operations. Traditionally, UAVs utilize exhaustive lawnmower exploration patterns to locate stranded survivors, without any information regarding the survivor's whereabouts. In real life disaster scenarios however, on-ground observers provide valuable information to the rescue effort, such as the survivor’s last known location and heading. In earlier work, a Weight Based Exploration (WBE) model, which utilizes this information to generate a prioritized list of waypoints to aid the UAV in its search mission, was proposed. This approach was shown to be effective for a single UAV locating a single survivor. In this paper, we extend the WBE model to a team of UAVs locating multiple survivors. The model initially partitions the search environment amongst the UAVs using Voronoi cells. The UAVs then utilize the WBE model to locate survivors in their partitions. We test this model with varying survivor locations and headings. We demonstrate the scalability of the model developed by testing the model with aerial teams comprising several UAVs.

\end{abstract}


\end{frontmatter}


\section{Introduction}
\label{S:1}

Floods cause damage to communities with loss to human life and infrastructure. In recent years, several technological solutions, in areas such as dynamic routing of resources \cite{Kashyap2019} and monitoring of fast flowing waters \cite{perks2016advances}, have been proposed to mitigate the effects of such natural disasters \cite{Avalanche}, \cite{HumanBodyDetection}. Given their ability to rapidly traverse large areas affected by fast moving flood waters, UAVs are becoming increasingly popular as an aid in rescue and monitoring efforts \cite{Sensors2017}, \cite{Ravichandran2019}, \cite{shetty2020implementation}. During rescue efforts, UAVs traditionally use the exhaustive lawnmower search pattern to explore a given area for survivors. Albeit effective, this model requires a significant amount of time to completely survey a given area and locate survivors. Such models do not leverage valuable observations made by on-ground rescue personnel, such as the survivor's last known position and heading. In \cite{Ravichandran2019}, a Weight-Based Exploration (WBE) model, which leverages valuable information regarding the survivor, as conveyed by on-ground observers, was described to generate a prioritized list of waypoints for exploration to aid the UAV in its search for survivors. This WBE model reduces the time taken by the UAV to locate survivors by nearly 215\% in Gazebo simulations and 75\% in physical testing, in comparison to the lawnmower model, using an off-the-shelf UAV \cite{shetty2020implementation}.

\medskip

In order to survey large areas effectively in a short span of time, the use of UAV teams has been found to offer a possible viable solution \cite{Hildmann_2019}. In this paper, the authors extend the WBE model to a team of UAVs, that collectively explore a given area for multiple flood survivors.

\medskip

The paper is organized as follows: The Weight-Based Exploration (WBE) model is first described in Section \ref{S:2}. In Section \ref{S:3}, the use of WBE for multi-UAV teams is described, followed by a description of the simulation environment and the results of the simulations in Section \ref{S:4}. The paper is concluded in Section \ref{S:5}, with a brief account of the paper's contribution and suggestions for further work.

\section{The Basic Weight-Based Exploration Model and Extensions}
\label{S:2}

\begin{figure}[h!]
    \centering
    \fbox{\includegraphics[scale=0.30]{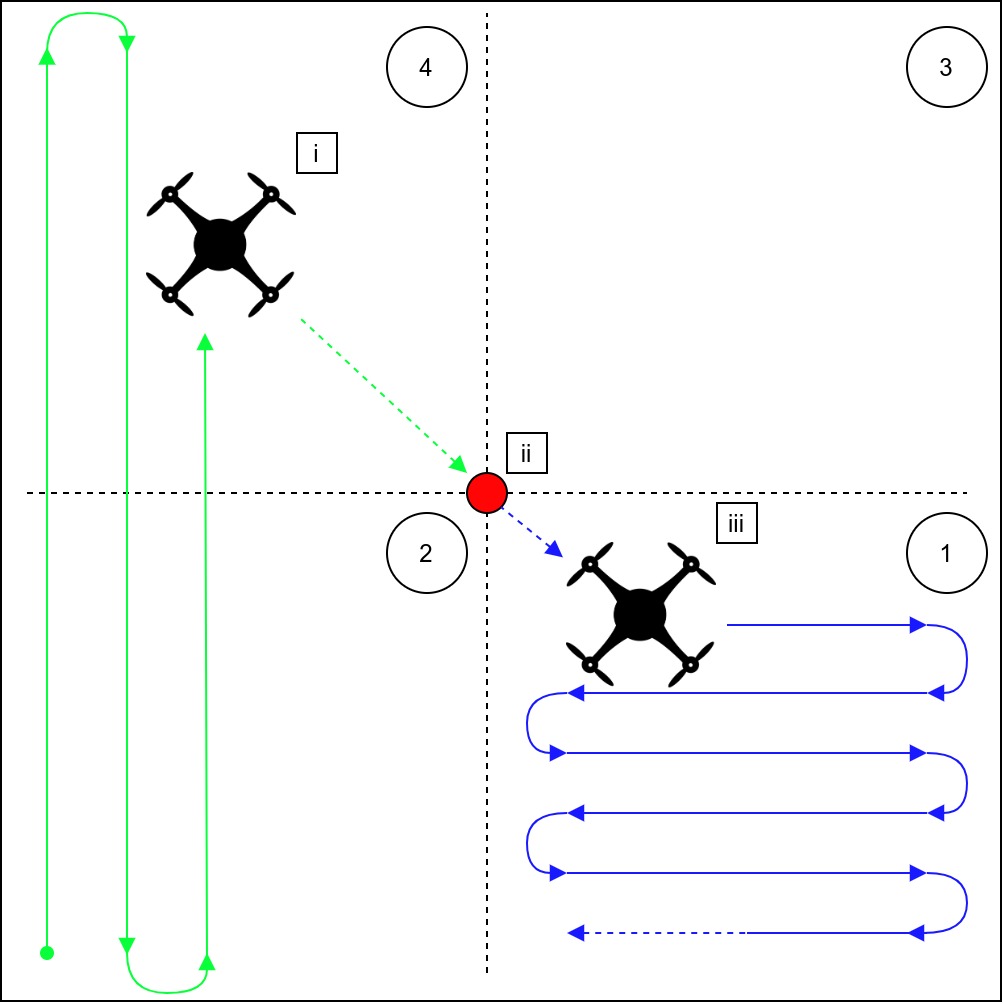}}
    \caption{Diagrammatic Representation of Weight-Based Exploration for a single UAV}
    \label{fig:WeightSingle}
\end{figure}

The Weight-Based Exploration model, first described in \cite{Ravichandran2019}, leverages the last known position and heading of a survivor, as reported by an on-ground observer, to generate a prioritized list of waypoints for exploration, which enables a UAV to locate the survivor in a short time-span. Figure \ref{fig:WeightSingle} is a diagrammatic representation of this model for a single UAV case.

\setlength{\fboxsep}{1.3pt}%
\setlength{\fboxrule}{1pt}%

In the WBE model, the coordinates lying along the direction of the survivor's heading receives the highest "weight" and therefore the highest priority of exploration. The second highest priority of exploration is assigned to the quadrant on the left of the survivor's heading and the quadrant on the right receives the third highest priority. The priorities assigned to these two quadrant are interchangeable. The direction opposite to the survivor's heading, is assigned the least priority of exploration. This differential assignment of weights to the environment coordinates, based on their proximity to the survivor's location and heading, allows us to arrive at a list of waypoints, where the highest weighted waypoints represent the most probable location where the survivor might be present.

In Figure \ref{fig:WeightSingle}, the UAV initially performs the lawnmower exploration (path in green, marked as i) to locate survivors. Once it receives information regarding a survivor's last known position and heading from the observer (red circle, marked as ii), it breaks free from the lawnmower model and initiates the WBE model (path in blue, marked as iii), starting from the survivor's last known position.

\begin{table}[h!]
\caption{Parameters of Monte-Carlo Simulations \cite{Ravichandran2019}}
\centering
\begin{tabular}{|l|l|}
\hline
\textbf{Parameter}     & \textbf{Value}    \\ \hline
Environment Size       & 600 m x 600 m     \\ \hline
Survivor Speed         & 0.6 m/s and 2 m/s \\ \hline
UAV Max Speed          & 12 m/s            \\ \hline
UAV Search Radius      & 18 m x 18 m       \\ \hline
UAV Camera FOV         & \ang{45}               \\ \hline
UAV Flight Time        & 30 minutes        \\ \hline
Height of UAV          & 9 m               \\ \hline
Observer Search Radius & 30 m              \\ \hline
Number of observers    & 30                \\ \hline
Observer Positions     & Randomized        \\ \hline
\end{tabular}
\label{tab:MonteCarlo}
\end{table}

In \cite{Ravichandran2019}, Monte-Carlo simulations were used to assess the performance of the WBE model against the lawnmower model with the 12 parameters given in Table \ref{tab:MonteCarlo}. These simulations were run on \texttt{MATLAB}.

\setlength{\fboxsep}{1.3pt}%
\setlength{\fboxrule}{1pt}%
\begin{figure}[h!]
    \centering
    \subfloat{\fbox{\includegraphics[scale=0.253]{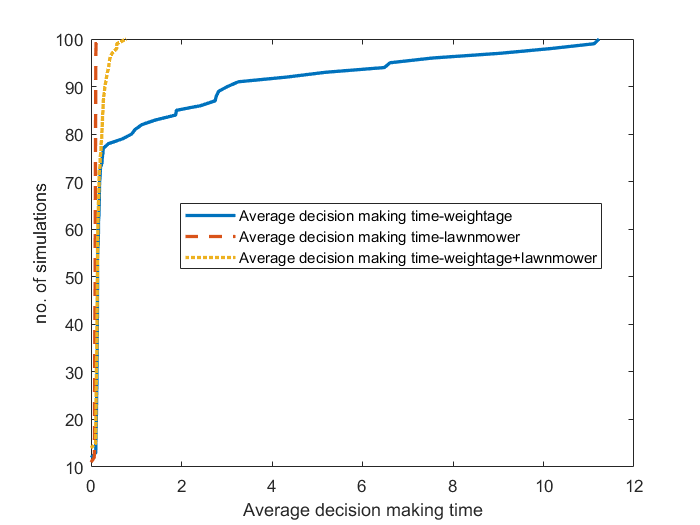}}}
    \qquad
    \subfloat{\fbox{\includegraphics[scale=0.347]{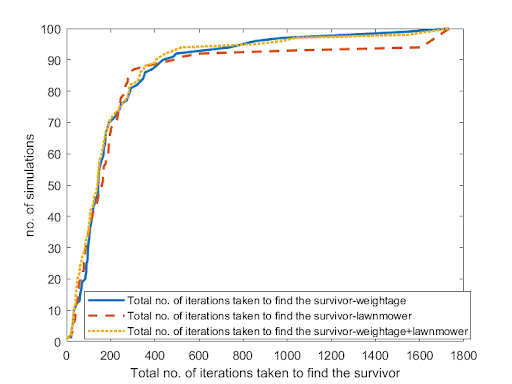}}}
    \captionsetup{justification=centering}
    \caption{Number of Simulations vs Average Decision Making Time and \\ Number of Simulations vs Total Iterations to Locate the Survivor \cite{Ravichandran2019}}
    \label{fig:Monte}
\end{figure}

The results from the Monte Carlo simulations are presented in Figure \ref{fig:Monte}. Even though the decision making time of the WBE model is greater than the lawnmower model, the WBE model requires fewer number of iterations to locate the survivor. In \cite{shetty2020implementation}, this WBE model was further developed by devising a set of equations for assigning the weights to the environment coordinates for prioritization. These equations are presented below:

\begin{align}
\emph{W\textsubscript{1}}&=(\emph{W\textsubscript{4}} N^3) + N^2 + N + 1 \\
\emph{W\textsubscript{2}}&=(\emph{W\textsubscript{1}} - 1) / N \\
\emph{W\textsubscript{3}}&=(\emph{W\textsubscript{1}} - N - 1) / N^2 \\
\emph{W\textsubscript{5}}&=(\emph{W\textsubscript{1}} N) + 1
\end{align}

Here, $N$ is the number of iterations required to move from the survivor's last known position to the boundary of the search environment, $W\textsubscript{1}$ is the weight assigned to the coordinates lying along the direction of the survivor's heading, $W\textsubscript{2}$ and $W\textsubscript{3}$ are the weights assigned to the left and right of the survivor heading, and $W\textsubscript{4}$ is the weight allotted to coordinates lying opposite to the survivor's heading. The weight $W\textsubscript{4}$ is assigned the least non-negative integer value of 1. $W\textsubscript{5}$ is given the highest weight value, assigned to the last known position of the survivor. These equations prevent the erratic movement of the UAV from one quadrant to another during the search operation. This erratic path may arise due to conflicting priorities of exploration that was observed in earlier variants of the WBE model.

\medskip

\begin{figure}
    \centering
    \subfloat{\fbox{\includegraphics[scale=0.38]{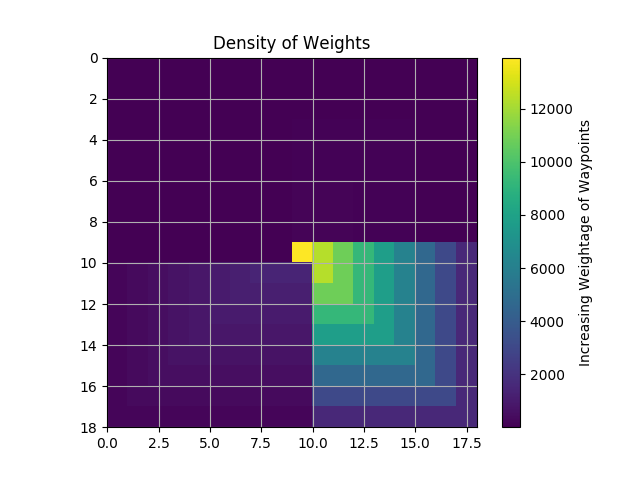}}}
    \qquad
    \subfloat{\fbox{\includegraphics[scale=0.38]{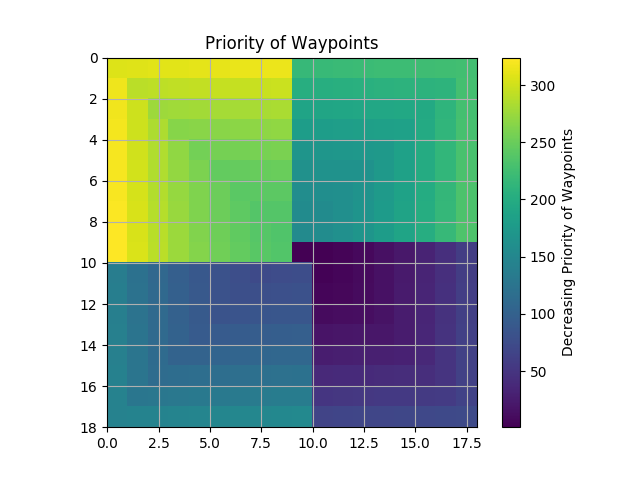}}}
    \caption{Density of Weights and Priority of Exploration Maps\cite{shetty2020implementation}}
    \label{fig:WeightageMap}
\end{figure}

The equations given above were used to generate waypoints in an example 18 m x 18 m environment, where the survivor was last reported at the 9 m x 9 m mark, moving in the south-east direction. Due to the differential assignment of weights to the coordinates, a color gradient can be observed across the quadrants in the Density of Weight Map, which translates to a variation in exploration priority in the Priority of Waypoints Map in Figure \ref{fig:WeightageMap}. This improved model was then used to locate survivors in multiple simulated environments and survivor configurations using Gazebo and ROS \cite{shetty2020implementation}. The resulting survivor search-times for the WBE (\textbf{\emph{T\textsubscript{W}}}) and lawnmower (\textbf{\emph{T\textsubscript{L}}}) models are presented in Table \ref{tab:SearchTime}. The WBE model clearly outperforms the lawnmower model. The WBE model was then physically implemented on an off-the-shelf UAV, running a \texttt{MAVROS} node on a Pixhawk 1 autopilot. The survivor search times of the WBE model were compared to that of a standard lawnmower model, in a 10 m x 10 m area with the survivor located at the the 5 m x 5 m mark.  While the lawnmower model required a total of 261 seconds to locate the survivor, the WBE model took 149 seconds, an improvement of 75\% \cite{shetty2020implementation}.

\setlength{\fboxsep}{1.3pt}%
\setlength{\fboxrule}{1pt}%

\begin{table}[h]
\caption{Search Times for Lawn-Mower and Weight-Based Exploration \cite{shetty2020implementation}}
\centering
\begin{tabular}{|l|l|l|l|}
\hline
\textbf{Environment Size} ($m^2$)    & \textbf{\emph{V\textsubscript{s}}} (m/s)     & \textbf{\emph{T\textsubscript{L}}} (s) & \textbf{\emph{T\textsubscript{W}}} (s)\\ \hline
18  x  18          & 0.6  & 624    & 173\\ \hline
20  x  20          & 0.6  & 669    & 213\\ \hline
18  x  18          & 0.3  & 600    & 63\\ \hline
20  x  20          & 0.3  & 663    & 66\\ \hline
\end{tabular}
\label{tab:SearchTime}
\end{table}

Given the improvement in performance of the WBE model, we adapted the WBE model to run on a team of UAVs, involved in locating multiple survivors in the given environment.

\section{Weight-Based Exploration for Multi-UAV Teams}
\label{S:3}

The exploration model for multi-UAV teams comprises two modules: 1. Voronoi Partitioning and 2. Weight-Based Exploration Model.

\subsection{Voronoi Partitioning}

Before the search operation begins, the entire environment space is divided amongst the UAVs using the Voronoi Partitioning algorithm, presented below  (Equation \ref{eq:VoronoiEquation}). The starting locations of the UAVs are used as generator points to create these partitions.

\begin{equation}\label{eq:VoronoiEquation}
    R_\textsubscript{k} = \{x \in X \: | \: d(x, k) \leq \: d(x, j) \: \forall \: k \neq j\}
\end{equation}

\setlength{\fboxsep}{1.3pt}%
\setlength{\fboxrule}{1pt}%
\begin{figure}[h!]
    \centering
    \fbox{\includegraphics[scale=0.34]{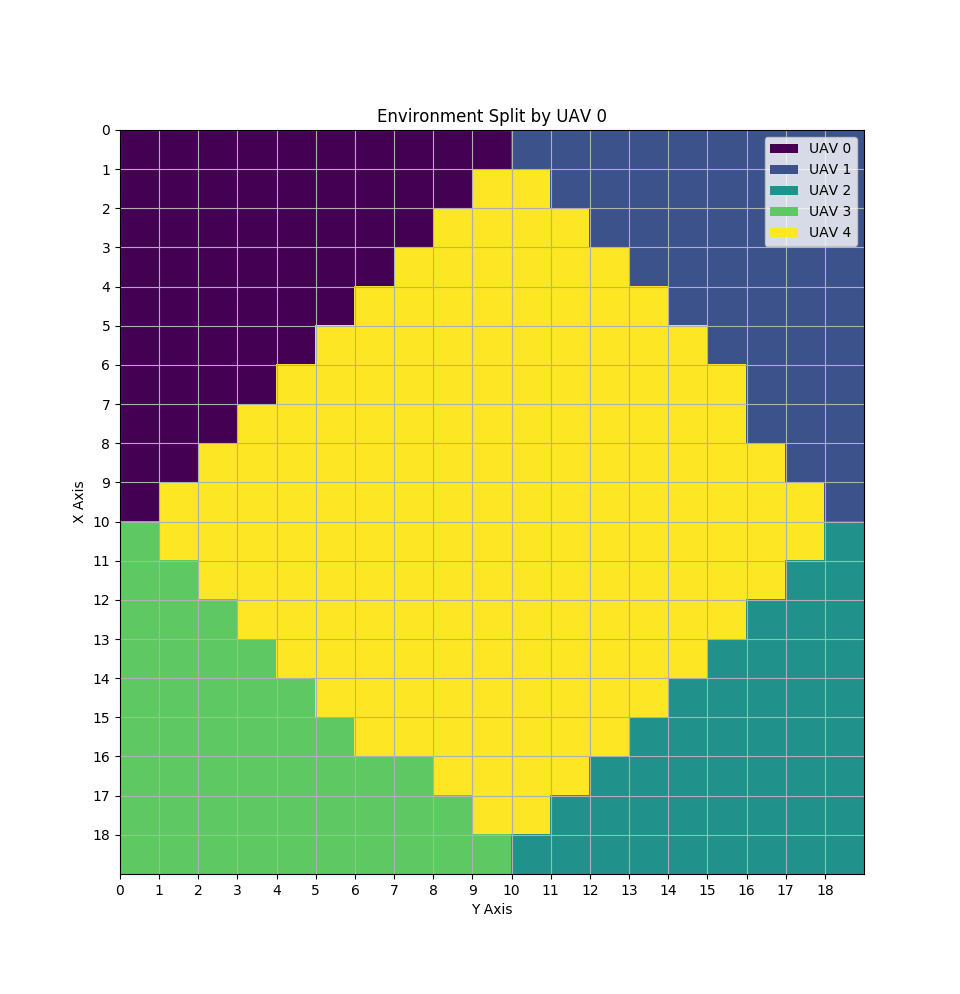}}
    \caption{Voronoi Partition of the Environment}
    \label{fig:VoronoiSplit}
\end{figure}

In Equation \ref{eq:VoronoiEquation}, $R_\textsubscript{k}$ represents the subset of all points, $x$ belonging to a space $X$, such that the distance to point $x$ from $k$, $d(x, k)$, is less than the distance, $d(x, j)$, to any other point $j$, where $j \neq k$.

\medskip

Once the environment is partitioned, the UAVs explore only their respective partitions, initially using the lawnmower model, and then the WBE model if triggered by an observer. The Voronoi Partitions for a 20 m x 20 m environment amongst a team of 5 UAVs is shown in Figure \ref{fig:VoronoiSplit}. In this example, the starting positions of the 5 UAVs were (0, 0), (0, 19), (19, 0), (19, 19) and (9, 9). The dimensions presented are for illustrative purposes and the speeds of the UAVs are compatible to the dimension of the exploration space. These numbers can be easily scaled up to represent real-life scenarios.

\subsection{Weight-Based Exploration Model in Partitioned Search Space}

\setlength{\fboxsep}{1.3pt}%
\setlength{\fboxrule}{1pt}%
\begin{figure}[h!]
    \centering
    \fbox{\includegraphics[scale=0.3]{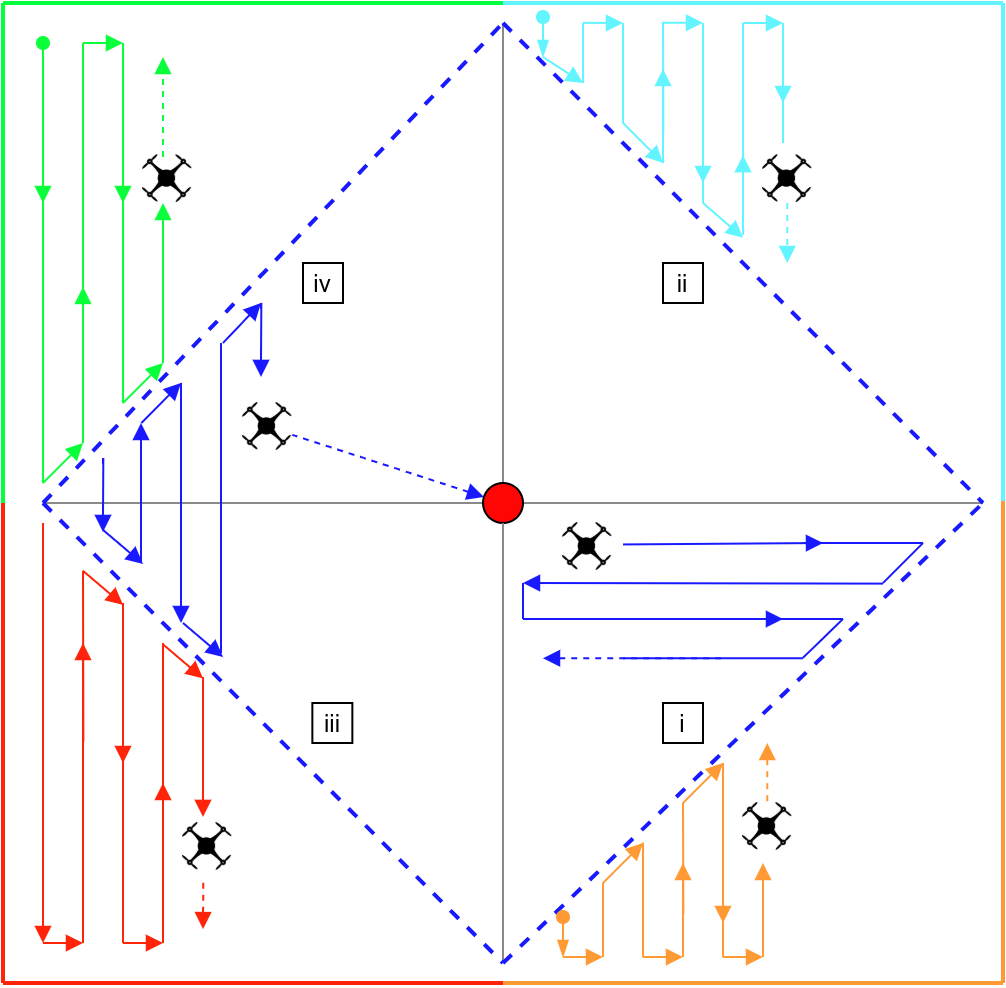}}
    \caption{Weight-Based Exploration with Multiple UAVs}
    \label{fig:VoronoiDiagram}
\end{figure}

Once the partitions have been generated, the UAVs commence their search for the survivors. The diagrammatic representation is shown in Figure \ref{fig:VoronoiDiagram}, where a team of 5 UAVs are participating in the search for 5 survivors in their respective colored partitions. The UAV in the blue partition, is triggered by an on-ground observer (red circle), upon locating a survivor in its vicinity. This UAV breaks off from the initial lawnmower model and heads to the highest priority quadrant within its partition marked 'i' and initiates the WBE model. The UAV searches for the survivor only in its partition, while the rest of the agents continue their lawnmower exploration, independently, unless triggered by their observers.

\medskip

In the following section, we describe the simulation environment and the results of running the WBE model on multiple
UAVs with varying survivor configurations.

\section{Simulation Environment and Results}
\label{S:4}

We simulate the model on ROS using the Gazebo simulator, on a UAV team comprising of 3DR Iris Quadcopters, running a \texttt{MAVROS} node on-board a Pixhawk 1 autopilot board. This configuration of the Quadcopter and autopilot is selected as it closely resembles the physical hardware on which these models will eventually be run.

\setlength{\fboxsep}{1.3pt}%
\setlength{\fboxrule}{1pt}%
\begin{figure}[h!]
    \centering
    \subfloat{\fbox{\includegraphics[scale=0.215]{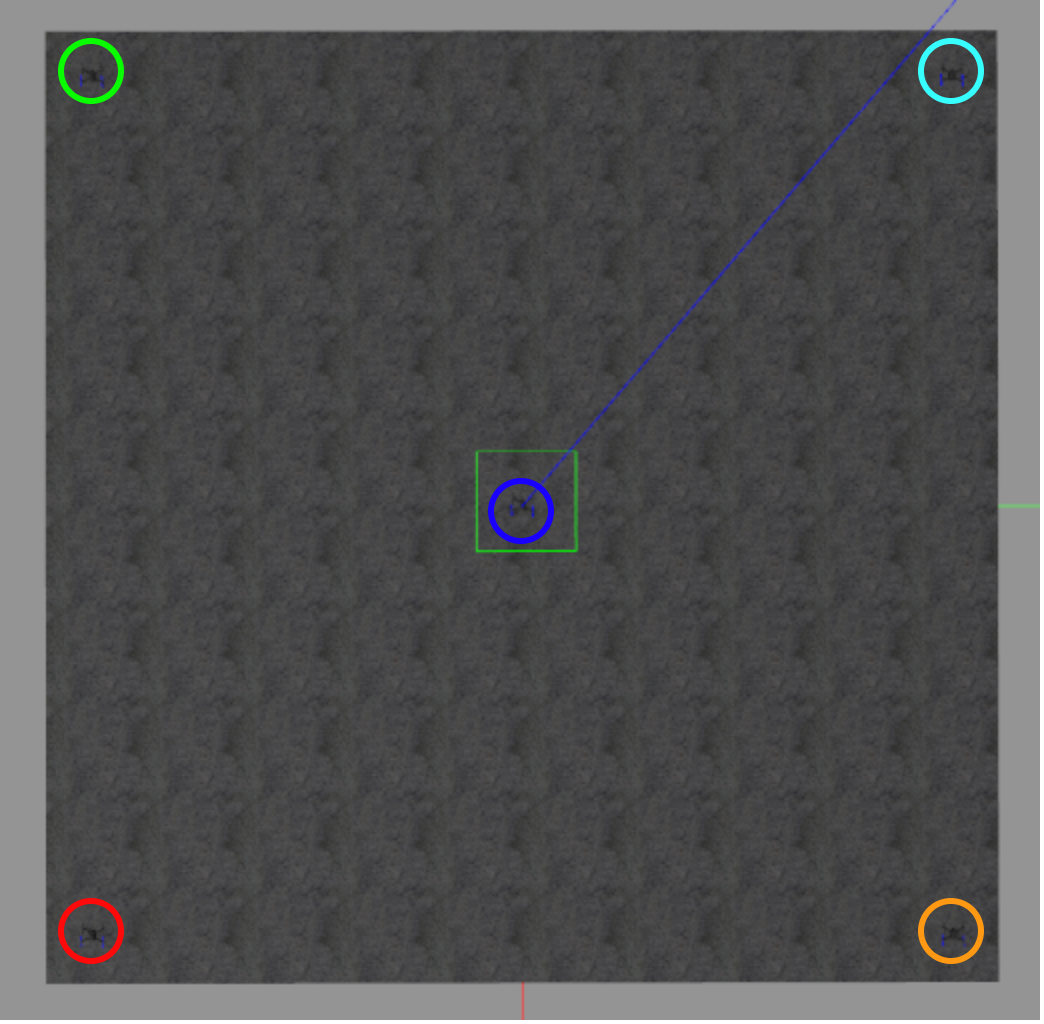}}}
    \qquad
    \subfloat{\fbox{\includegraphics[scale=0.181]{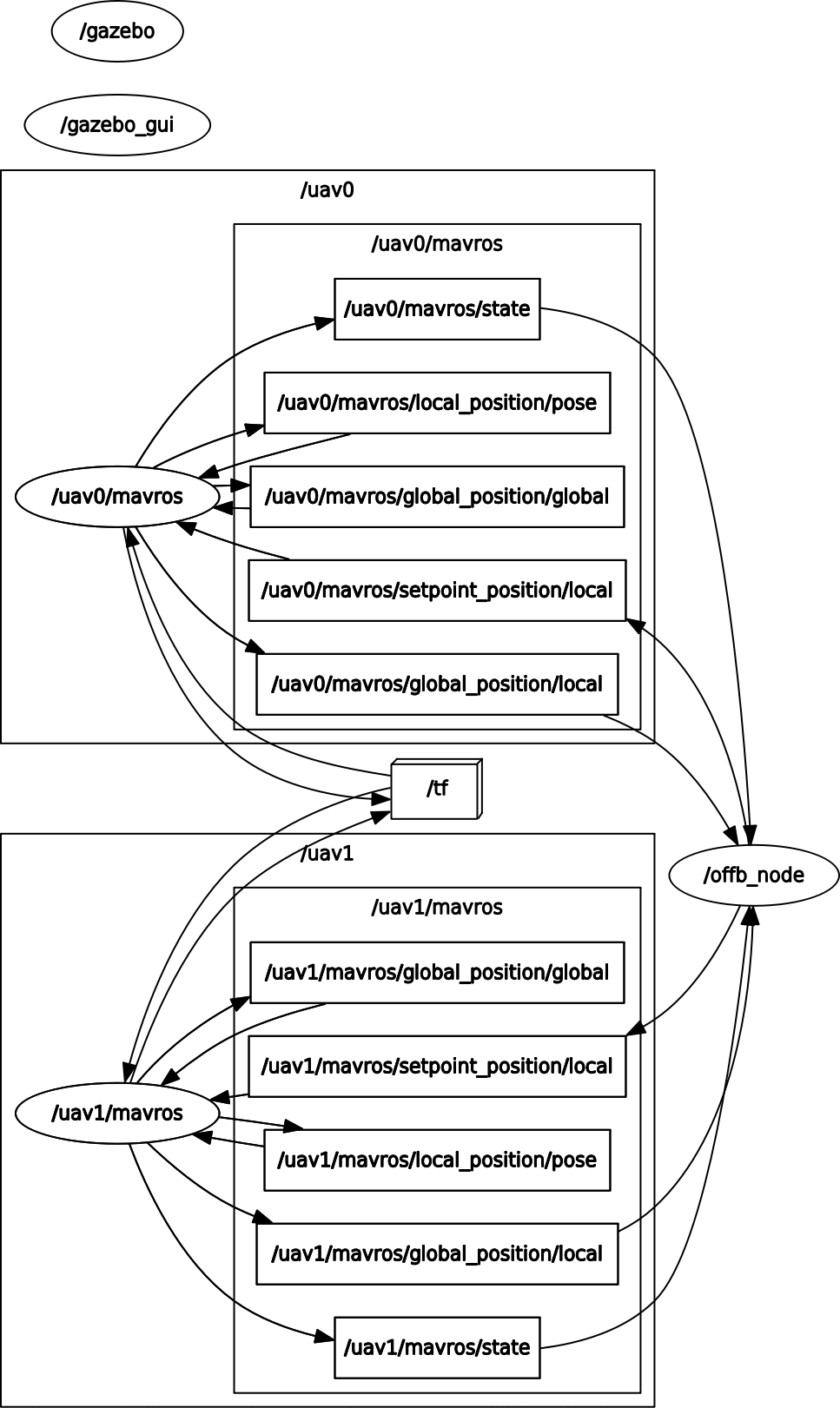}}}
    \caption{ROS Simulation Environment and \texttt{rqt\_graph}}
    \label{fig:ROSEnvironment}
\end{figure}

Figure \ref{fig:ROSEnvironment} shows the Gazebo environment with an example team of 5 UAVs that will be performing the survivor search. Figure \ref{fig:ROSEnvironment} also shows the \texttt{rqt\_graph} for a 2 UAV team simulation, with the various UAV and survivor topics and nodes generated during the simulation. A ROS \texttt{observer\_node} alerts the UAVs of the presence of a survivor in their vicinity.

Next, we present the results of the WBE model, with varying team configurations, ranging from 3 to 5 UAVs. The environment surveyed by the UAVs in each example is 20 m x 20 m. The survivor in each case moves linearly with a velocity of 0.5 m/s in a prescribed direction. The UAVs investigate the environment from a height of 2 m, with a field-of-view of \ang{45}.

\medskip

The 3-dimensional trajectory plot in each result section is accompanied by an X-Y projection for a better view of the trajectories. Detailed variations in the positions of both the UAVs and survivors are presented as well. The data is plotted using the \texttt{matplotlib} \cite{matplotlib} library for Python.

\subsection{Three UAV Team}

\setlength{\fboxsep}{1.3pt}%
\setlength{\fboxrule}{1pt}%
\begin{figure}[h!]
    \centering
    \subfloat{\fbox{\includegraphics[scale=0.22]{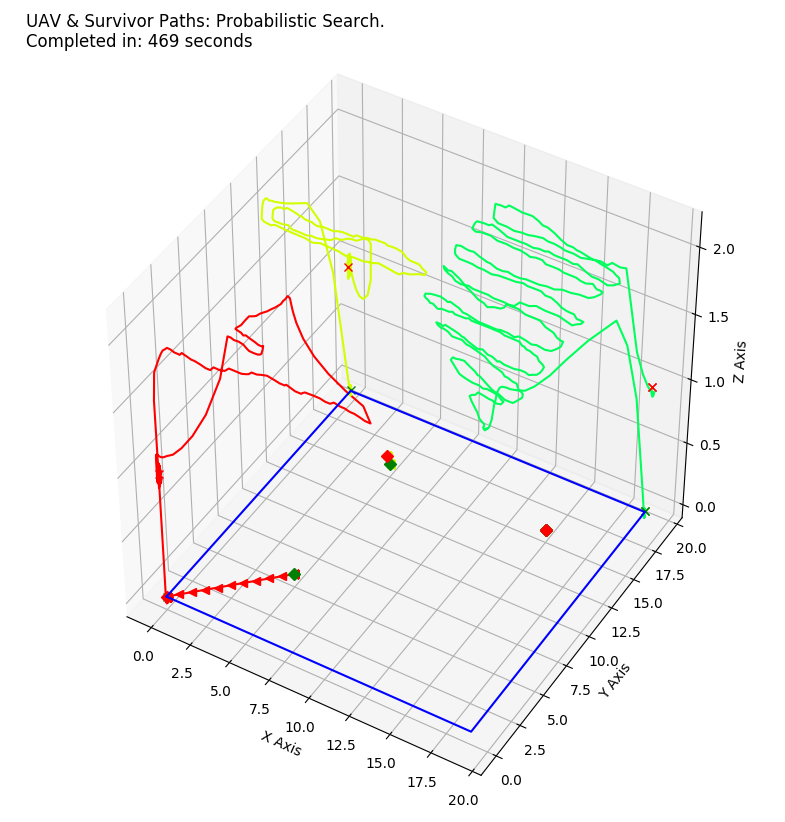}}}\hspace*{-1.5em}
    \qquad
    \subfloat{\fbox{\includegraphics[scale=0.6]{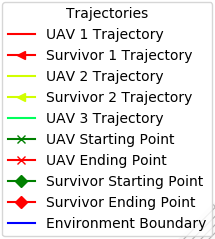}}}\hspace*{-1.5em}
    \qquad
    \subfloat{\fbox{\includegraphics[scale=0.22]{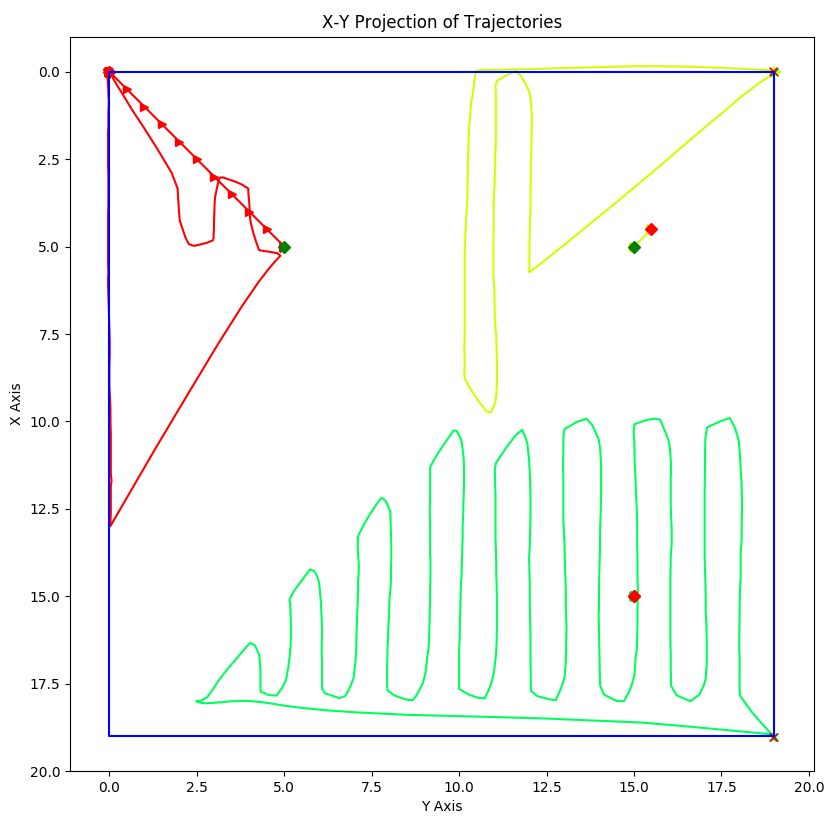}}}
    \caption{Trajectories and X-Y Projection for 3 UAVs and Survivors}
    \label{fig:Result2Trajectories}%
\end{figure}

\setlength{\fboxsep}{1.3pt}%
\setlength{\fboxrule}{1pt}%
\begin{figure}[h!]
    \centering
    \fbox{\includegraphics[scale=0.305]{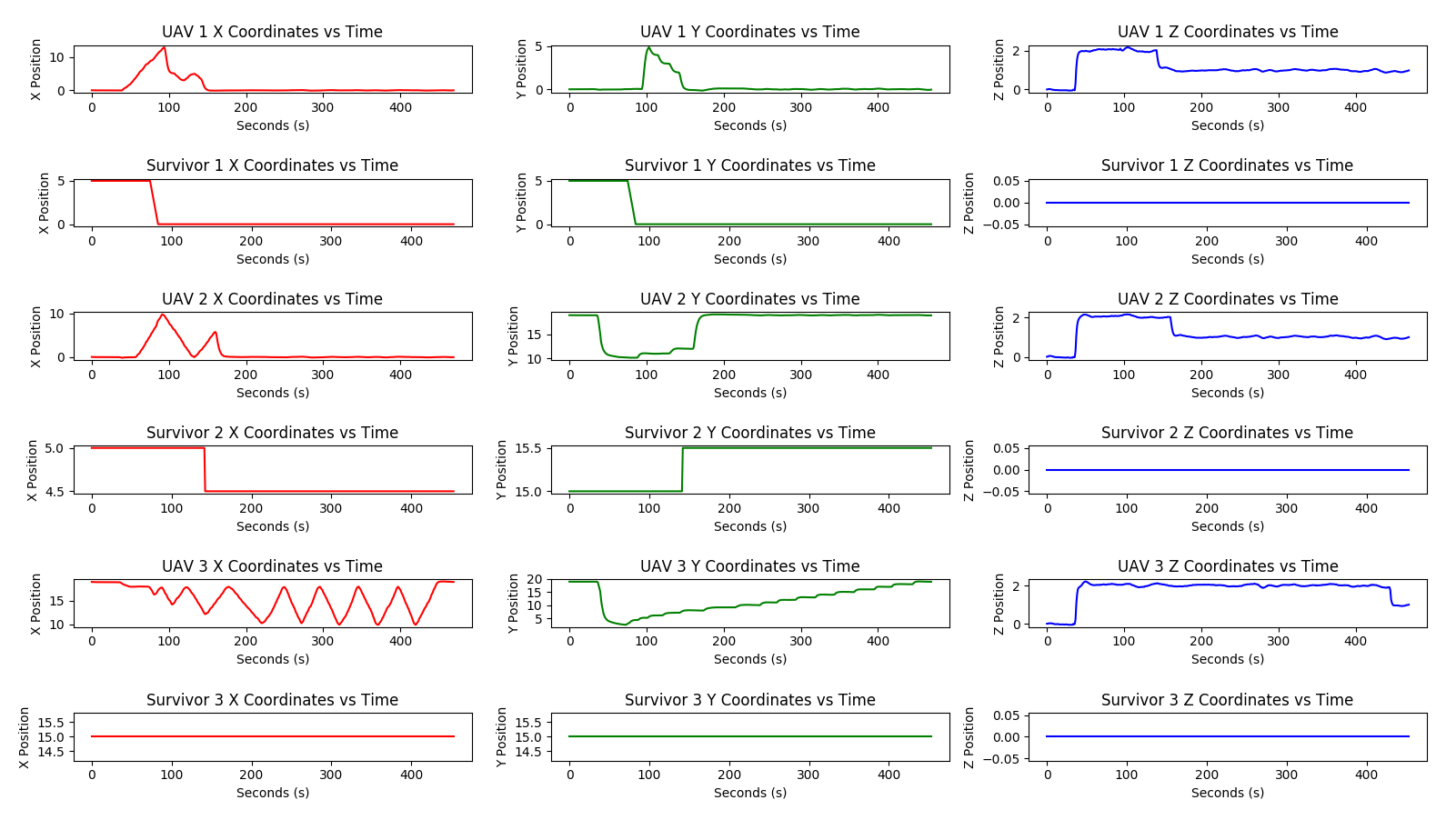}}
    \caption{Variation in Position for 3 UAVs and Survivors}
    \label{fig:Result2Info}
\end{figure}

We initialize 3 UAVs, located at (0, 0), (0, 19) and (19, 19). The survivors are located at (5, 5), (5, 15) and (15, 15). As seen in Figure \ref{fig:Result2Trajectories}, UAV 1 and 2 are triggered by their observers due to the presence of survivors in their vicinity. This causes UAV 1 and 2 to break off from the lawnmower exploration and head to their survivors last known location and and initiate the WBE model.

Due to the modular nature of the model, UAV 3 continues to perform the lawnmower exploration, whereas UAVs 1 and 2 execute the WBE model until they locate their survivor. Once the UAVs locate their survivor in their field-of-view, they head back to their starting positions. The variation in the positions for all the UAVs and survivors are presented in Figure \ref{fig:Result2Info}.

\subsection{Four UAV Team}

\setlength{\fboxsep}{1.3pt}%
\setlength{\fboxrule}{1pt}%
\begin{figure}[h!]
    \centering
    \subfloat{\fbox{\includegraphics[scale=0.225]{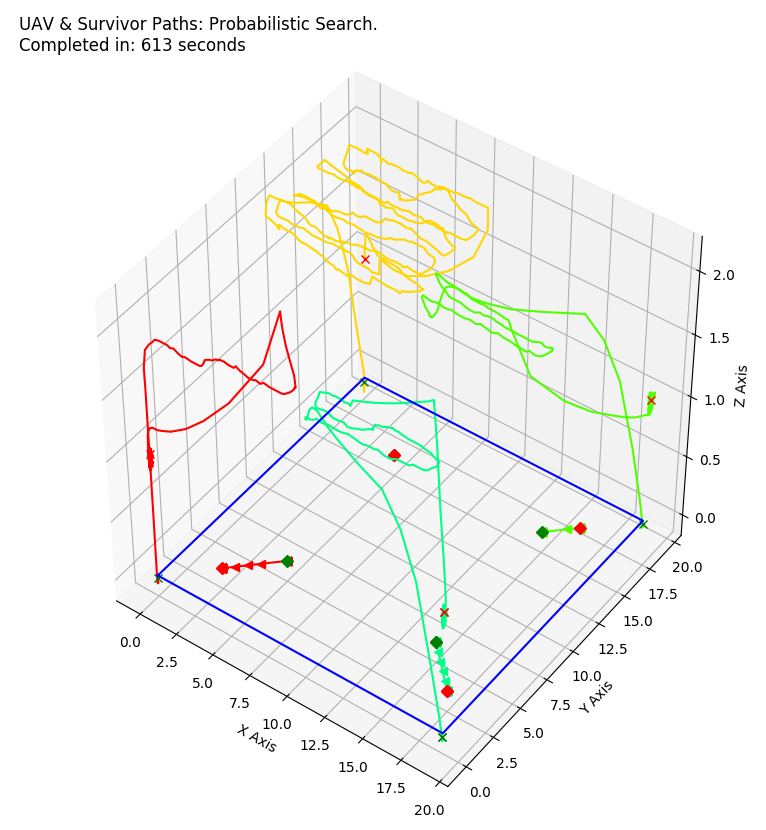}}}\hspace*{-1.5em}
    \qquad
    \subfloat{\fbox{\includegraphics[scale=0.6]{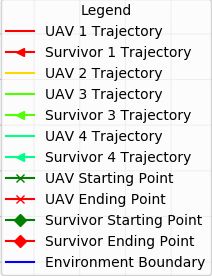}}}\hspace*{-1.5em}
    \qquad
    \subfloat{\fbox{\includegraphics[scale=0.225]{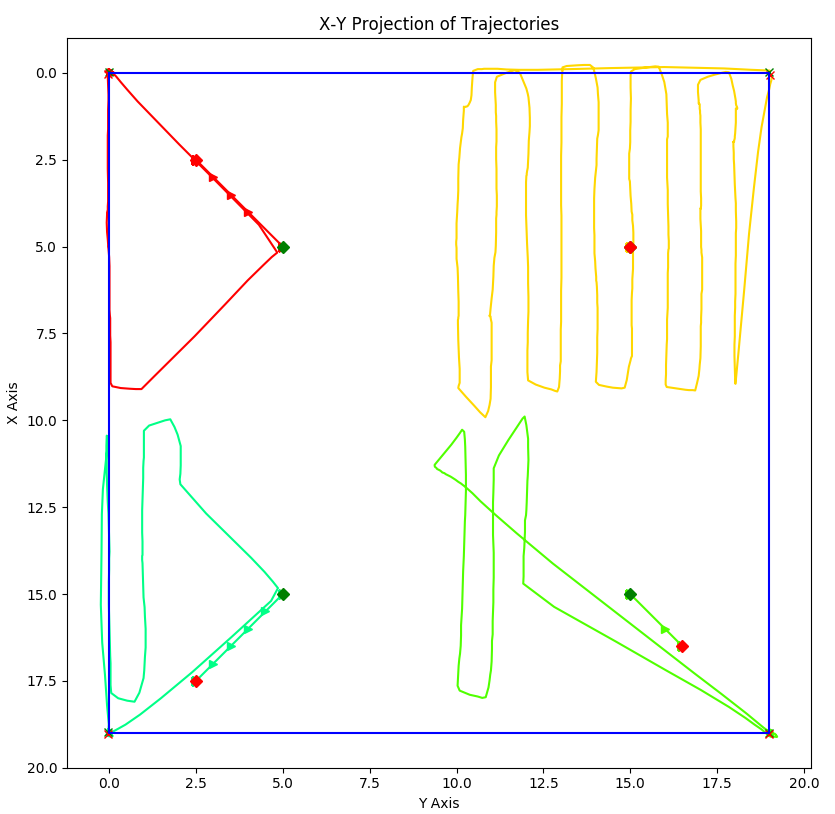}}}
    \caption{Trajectories and X-Y Projection for 4 UAVs and Survivors}%
    \label{fig:Result3Trajectories}%
\end{figure}

\setlength{\fboxsep}{1.3pt}%
\setlength{\fboxrule}{1pt}%
\begin{figure}[h!]
    \centering
    \fbox{\includegraphics[scale=0.297]{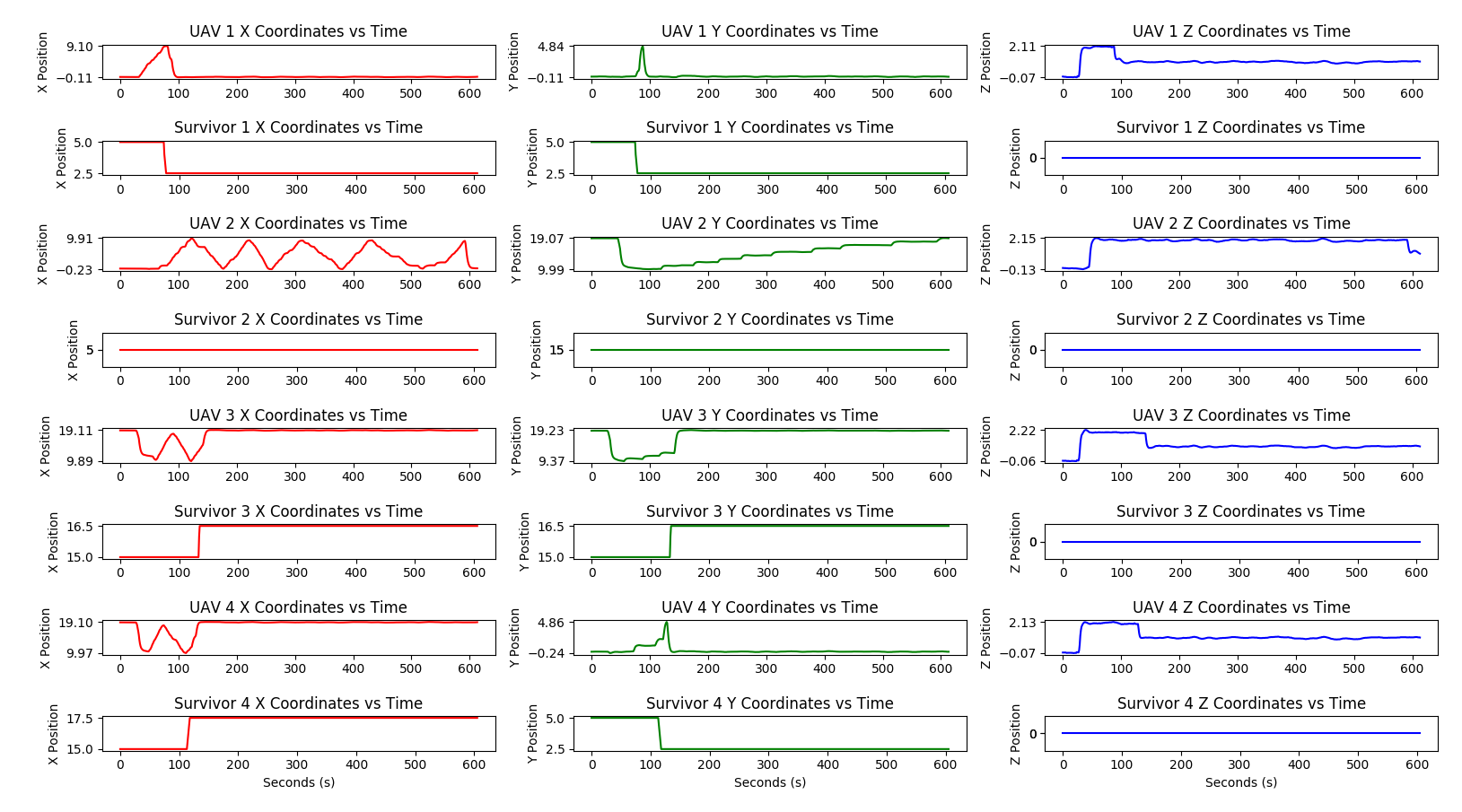}}
    \caption{Variation in Position for 4 UAVs and Survivors}
    \label{fig:Result3Info}
\end{figure}

Figure \ref{fig:Result3Trajectories} presents an aerial team with 4 UAVs, starting from (0, 0), (0, 19), (19, 19) and (19, 0). Their corresponding survivors are located at (5, 5), (5, 15), (15, 15) and (15, 5). Observers alert UAVs 1, 3 and 4 of survivor's in their partitions. The UAVs initiate the weight-based exploration to locate their moving survivors. As presented in earlier sections, the UAVs explore only their own partitions.

UAV 2 continues with the lawnmower exploration, independent of the other UAVs, since its observer did not detect any survivor in its vicinity. The variations in the UAV and survivor positions during the simulation are presented in Figure \ref{fig:Result3Info}.

\subsection{Five UAV Team}

\setlength{\fboxsep}{1.3pt}%
\setlength{\fboxrule}{1pt}%
\begin{figure}[h!]
    \centering
    \subfloat{\fbox{\includegraphics[scale=0.22]{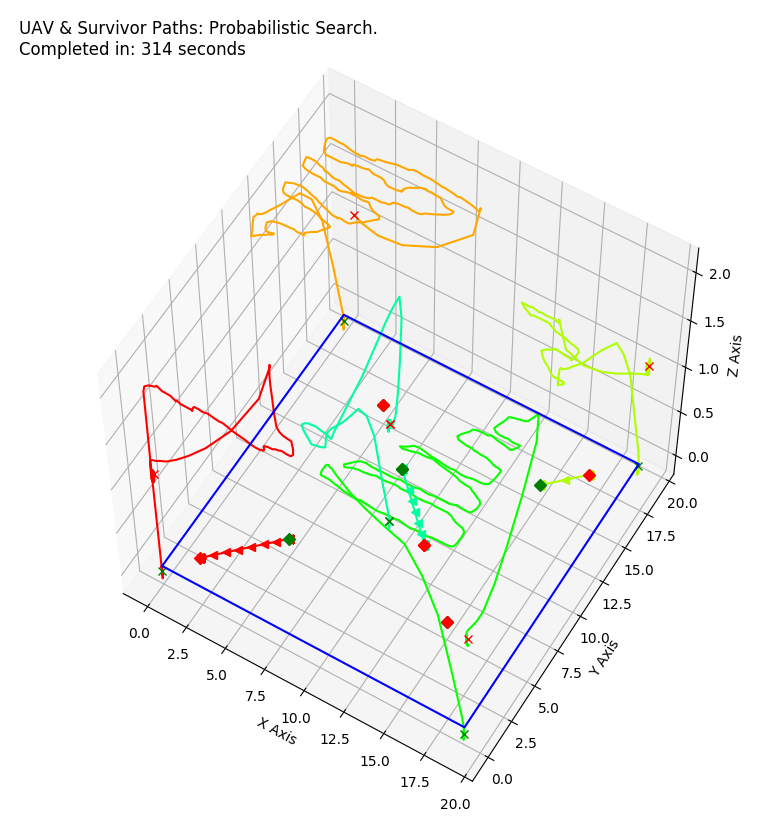}}}\hspace*{-1.5em}
    \qquad
    \subfloat{\fbox{\includegraphics[scale=0.5]{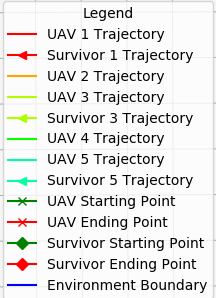}}}\hspace*{-1.5em}
    \qquad
    \subfloat{\fbox{\includegraphics[scale=0.225]{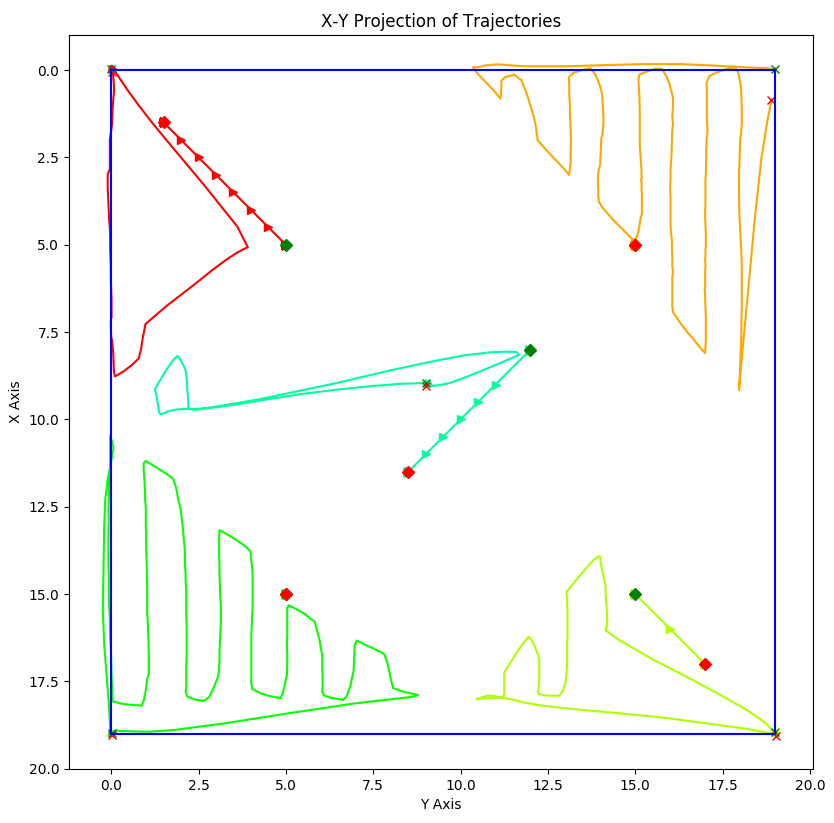}}}
    \caption{Trajectories and X-Y Projection for 5 UAVs and Survivors}%
    \label{fig:Result4Trajectories}%
\end{figure}

\setlength{\fboxsep}{1.3pt}%
\setlength{\fboxrule}{1pt}%
\begin{figure}[h!]
    \centering
    \fbox{\includegraphics[scale=0.280]{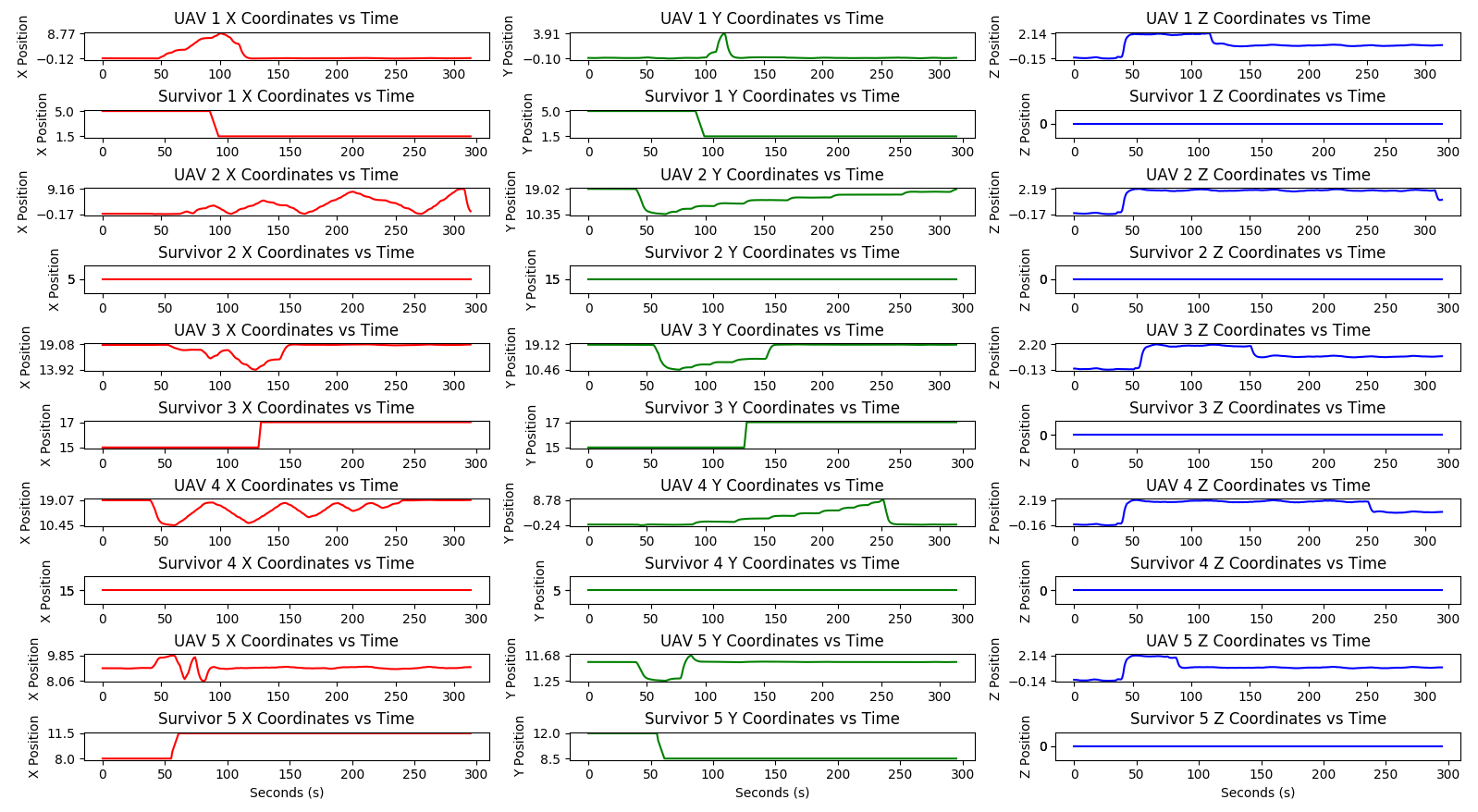}}
    \caption{Variation in Position for 5 UAVs and Survivors}
    \label{fig:Result4Info}
\end{figure}

A 5 UAV team involved in the search for 5 survivors in a 20 m x 20 m environment is shown in Figure \ref{fig:Result4Trajectories} and Figure \ref{fig:Result4Info}. The UAVs start from (0, 0), (0, 19), (19, 19), (19, 0) and (9, 9) respectively, with the corresponding survivors located at (5, 5), (5, 15), (15, 15), (15, 5) and (8, 12).

Observers alert UAVs 1, 3 and 5 about the presence of survivors in their partition. These UAVs initiate the weight-based exploration to locate their survivors. UAVs 2 and 4 continue exploring their partitions using the lawnmower model. Having successfully located their survivors using the WBE model, the UAVs return to their starting location.

In this section, we have successfully demonstrated the scalability of the weight-based model, when deployed on an aerial team comprising of multiple UAVs. In each case, the model initially partitions the environment amongst the UAVs, using their starting locations, and then initiates the exploration of the environment. If an observer alerts the UAV of the presence of a survivor in its vicinity, the UAV initiates the WBE model and begins locating the survivor in its partition. Given the modular nature of the model, the exploration model of a UAV remains unaffected by the exploration models of the other UAVs in the team.

\section{Conclusion and Future Work}
\label{S:5}
In this paper we presented a development to the previously described Weight-Based Exploration Model, to accommodate multi-UAV teams involved in search and rescue operations. Such UAV teams can aid rescue efforts and significantly reduce the time taken to investigate the given environment for survivors. In this paper, we have presented results for the WBE model executed on aerial teams with 3, 4 and 5 UAVs. However, with a sufficiently powerful system more UAVs can be simulated simultaneously.

\medskip

The model presented here explores the environment first by partitioning the search space amongst the UAVs who then locate survivors in their respective partitions. However, during physical implementation, UAVs are bound to have varying battery levels which will dictate the size of the areas that they will explore. We wish to develop the model so that factors such as battery levels and endurance can be taken into account while generating the partitions. Furthermore, agents can unexpectedly lose communication with each other or with the ground station. In such cases the model should allow for the re-partitioning of the environment amongst the remaining agents.

\medskip

In the future, we will integrate the WBE model with a computer-vision pipeline, trained on overhead images of humans to detect survivors. This would enable a truly end-to-end autonomous solution, where the WBE model will be used to locate survivors and then the computer-vision pipeline will identify them and alert the search effort. We aim to physically test the multi-UAV model integrated with the computer-vision pipeline, on off-the-shelf UAVs, with the eventual goal of deploying multiple exploring agents in a real-life flood scenario to aid rescue efforts.

\section*{Acknowledgements}
\label{S:6}

\noindent This work was partially supported by a Engineering and Physical Sciences Research Council - Global Challenges Research Fund (EPSRC-GCRF), UK, multi-institute grant (Grant Number: EP/P02839X/1).














\begin{thebibliography}{10}
\bibitem{Kashyap2019}
A. {Kashyap}, D. {Ghose}, P. P. {Menon}, P. B. {Sujit} and K. {Das},
\newblock{"UAV Aided Dynamic Routing of Resources in a Flood Scenario,"}
\newblock{International Conference on Unmanned Aircraft Systems (ICUAS), pp. 328-335, June 2019.}

\bibitem{perks2016advances}
T.M. {Perks}, A.J. {Russell} and A. R. {Large},
\newblock{"Advances in flash flood monitoring using unmanned aerial vehicles (UAVs),"}
\newblock {Hydrology \& Earth System Sciences, vol. 20, 2016.}

\bibitem{Avalanche}
M. {Silvagni}, A. {Tonoli}, E. {Zenerino} and M. {Chiaberge},
\newblock {"Multipurpose UAV for search and rescue operations in mountain avalanche events,"}
\newblock {Geomatics, Natural Hazards and Risk, pp. 18-33, 2007.}

\bibitem{HumanBodyDetection}
P. {Rudol} and P. {Doherty},
\newblock {"Human Body Detection and Geolocalization for UAV Search and Rescue Missions Using Color and Thermal Imagery,"}
\newblock{2008 IEEE Aerospace Conference, Big Sky, MT, pp. 1-8, 2008.}

\bibitem{Sensors2017}
 D. {Popescu}, L. {Ichim} and F.{Stoican},
\newblock{"Unmanned aerial vehicle systems for remote estimation of flooded areas based on complex image processing,"}
\newblock {Sensors, vol. 17, no. 3, pp. 446-470, 2017.}

\bibitem{Ravichandran2019}
R. {Ravichandran}, D. {Ghose} and K. {Das},
\newblock {"UAV Based Survivor Search during Floods,"}
\newblock {International Conference on Unmanned Aircraft Systems (ICUAS), pp. 1407-1415, 2019.}

\bibitem{shetty2020implementation}
S.J. {Shetty}, R. {Ravichandran}, L.A. {Tony}, N.S. {Abhinay}, K. {Das} and D. {Ghose},
\newblock{"Implementation of survivor detection strategies using drones,"}
\newblock{Unmanned Aerial Systems: Theoretical Foundation and Applications, in press.}

\bibitem{Hildmann_2019}
H. {Hildmann} and E. {Kovacs},
\newblock{"Review: Using Unmanned Aerial Vehicles (UAVs) as Mobile Sensing Platforms (MSPs) for Disaster Response, Civil Security and Public Safety,"}
\newblock {MDPI AG Drones, vol. 3, no. 3, pages 59 - 82, 2019.}

\bibitem{matplotlib}
J. D. {Hunter},
\newblock{"Matplotlib: A 2D graphics environment"},
\newblock {IEEE Computer Society Computing in Science \& Engineering, vol. 9, no. 3, pp. 90-95, 2007.}
\end{thebibliography}
\end{document}